# dashi: A Python library for Dataset Shift Characterization to Support Trustworthy AI Development and Deployment

David Fernández-Narro[1,*], Pablo Ferri[1], Ángel Sánchez-García[1], Juan M. García-Gómez[1] and Carlos Sáez[1]

[1]Biomedical Data Science Lab, Instituto Universitario de Tecnologías de la Información y Comunicaciones, Universitat Politècnica de Valéncia, Camino de Vera s/n, Valencia 46022, España

* Corresponding author

**Abstract**

**Background**

The Artificial Intelligence (AI) life cycle requires a thorough understanding of the underlying data dynamics for robust, safe and cost-effective AI development and use. Dataset shifts are defined as changes between train and test data distributions. Whether occurring over time (temporal) or across different sites (multi-source), they can severely degrade model performance and compromise data quality. This is particularly important in health AI, where the safety and fundamental rights of patients can be severely affected by uncontrolled shifts both at training and operational stages. While the theoretical foundations of covariate, prior, and concept shifts are well established, there is a lack of accessible and comprehensive software tools to perform their analysis.

**Results**

We introduce dashi, an open-source Python library designed for the exploration, quantification, and characterization of dataset shifts. dashi provides a dual approach: an unsupervised approach that leverages information geometry and non-parametric statistical manifolds to data variability characterization and analysis (e.g., Information Geometric Temporal plots and Multi-Source Variability metrics like Global Probabilistic Deviation and Source Probabilistic Outlyingness), and a supervised approach that quantifies and characterizes model performance degradation. Both unsupervised and supervised approaches work across user-defined temporal and domain/source batches. We demonstrate the utility of dashi on three simulated and real-world health AI case studies on gestational diabetes mellitus, COVID-19 and emergency medical dispatch. We also provide a table assembling different problems related to variability, how to detect them with dashi, their effect on AI and possible solutions.

**Conclusions**

By providing interactive visual analytics and variability metrics, dashi supports trustworthiness of AI life cycle stages enabling robust and safe machine learning pipelines through the assessment of data coherence and AI performance.

# 1. Introduction

## 1.1 Dataset shifts in AI

The traditional artificial intelligence (AI) lifecycle usually assumes a static and stable relationship between development and deployment scenarios. However, when integrating these systems into complex live domains, such as healthcare, this foundational assumption frequently collapses. The integration of AI into healthcare systems has accelerated rapidly, driven by the increasing availability of Electronic Health Records (EHRs) and the demand for advanced Clinical Decision Support Systems (CDSS) [1,2]. While predictive models offer significant potential to enhance diagnostic accuracy, optimize resource allocation, and personalize patient care, their deployment in live health environments introduce substantial risks [3]. Unlike controlled experimental settings, real-world data, like EHRs, is inherently dynamic and subject to continuous operational and populational evolution. This variability comprises the entire AI lifecycle: training on retrospective data fails to capture emerging clinical patterns, usage in live environments leads to degraded performance, and standard evaluation metrics provide a false sense of security by ignoring temporal and multi-source variability. These silent failures in predictive systems directly compromise patient safety, as clinicians may unknowingly rely on degraded, obsolete models for critical diagnostic or therapeutic decisions. Central to this operational paradigm is the concept of Trustworthy AI, which mandates that AI models remain reliable, transparent, and robust against the inevitable, often silent, instabilities of real-world data streams [1,4].

Dataset shifts occur when the joint probability distribution of patient features and clinical outcomes, $P(X,Y)$, changes between the environment where a model was trained and the clinical setting where it is actively deployed [5,6]. This statistical instability generally manifests in three distinct forms: *covariate shift*, where the distribution of input features $P(X)$ changes while the underlying conditional relationship remains stable (e.g., shifts in patient demographics or changing laboratory testing protocols); *prior shift* (or label shift), where the baseline prevalence of the outcome $P(Y)$ fluctuates while the conditional distribution of features given the outcome $P(X|Y)$ remains stable (e.g., seasonal variations in disease incidence); and *concept shift*, where the fundamental relationship between clinical features and the outcome $P(Y|X)$ evolves while the input feature distribution $P(X)$ remains stable (e.g., the introduction of a novel medical intervention that fundamentally alters disease progression). In clinical practice, these shifts are frequently driven by temporal dynamics, such as evolving clinical guidelines or abrupt epidemiological disruptions like the COVID-19 pandemic, as well as at model transportability, where multi-source spatial variations can create profound domain adaptation challenges when transferring models across different hospitals or services. The primary clinical risk associated with these phenomena is the occurrence of "silent failures." In these scenarios, the AI system continues to generate predictions without proper safety warnings [1,7], yet its underlying statistical assumptions have been violated. This model degradation can lead to poorly calibrated outputs, biased predictions, and ultimately, unsafe clinical decisions that can compromise patient care [8,9].

## 1.2 Limitations of current approaches

Throughout the clinical AI lifecycle, from initial training to deployment, rigorous dataset shift analysis is frequently bypassed. During model development, practitioners typically rely on traditional data quality assessments that prioritize static, univariate dimensions such as completeness, conformance, and basic summary statistics [10,11]. Subsequently, the evaluation and deployment phases fixate predominantly on retrospective predictive performance metrics. While standard quality checks are necessary for routine database maintenance, they are fundamentally ill-equipped to detect complex, multivariate distributional changes over time or across institutions. By neglecting proactive shift monitoring across all phases, this standard paradigm fails to identify "unknown unknowns", subtle yet highly impactful alterations in joint probability distributions that evade standard univariate alerts, ultimately leading to the silent degradation of clinical decision support systems [12,13].

Although the theoretical foundations of dataset shift and variability characterization have been extensively formalized in machine learning literature [6], there has been a historical lack of accessible, model-agnostic software tools designed to operationalize these advanced concepts. Translating abstract statistical changes into actionable insights for daily health MLOps procedures remains a significant bottleneck, leaving a critical gap between theoretical awareness and practical model monitoring.

### 1.3 The dashi library

In this work we introduce dashi, an open-source Python library designed specifically for analyzing and characterizing temporal and multi-source dataset shifts, overcoming the limitations in current dataset shifts characterization and monitoring frameworks. With the aim to bridge the gap between theoretical data drift detection and practical MLOps procedures, dashi equips AI developers with a robust dual approach: unsupervised and supervised analytical tools. The unsupervised framework provides a distribution-based, model-agnostic methodology that projects statistical distributions onto non-parametric statistical manifolds, allowing for the visual and quantitative delineation of shifts (Sáez et al., 2015; Sáez, Robles, & García-Gómez, 2017). Concurrently, the supervised approach offers a model-based characterization by automating the training and cross-batch evaluation of predictive models, such as Random Forests, thereby pinpointing the exact areas of temporal or spatial performance degradation. By comprehensively addressing covariate, prior, and concept shifts, the library helps identifying these effects and prevent silent model obsolescence in dynamic healthcare environments as well as help address data coherence as part of DQ assessment.

The dashi Python library is built upon the foundational work of the R package EHRTemporalVariability [16], advancing its core capabilities to meet the complex demands of AI. We have significantly extended its original unsupervised temporal methodology to support both conditional and multivariate dataset shift analyses. In addition, dashi introduces two completely novel components: an unsupervised module dedicated to multi-source characterization and validated stability metrics (Sáez, Robles, & García-Gómez, 2017), and a comprehensive supervised framework for tracking predictive performance degradation across varying data batches. The source code is freely available under an open-source license at https://github.com/bdslab-upv/dashi.

In the following sections, we first formalize the computational methodologies underlying the dashi framework. We then demonstrate its practical utility through applications on public and private cohorts, including a COVID-19 dataset, a Gestational Diabetes Mellitus (GDM) simulated dataset, and a Medical Call Incidents dataset compiled from the Health

Services Department of the Valencian Region, we illustrate how these tools operate in practice. Finally, we synthesize these findings to discuss their broader implications for the development, deployment and continuous monitoring of trustworthy health AI.

## 2. Methods and Implementation

dashi is architected as an open-source Python library designed to monitor all phases of health AI lifecycle. It is built with a modular, object-oriented structure that provides biomedical data scientists with a standardized pipeline for detecting and characterizing dataset shifts over time or across sources.

Regarding the shift analysis, the architecture of dashi is bifurcated into two complementary analytical modules. An unsupervised (data-centric) module and a supervised (model-centric) module:

- The unsupervised module is AI model-agnostic, i.e., it operates independently of any predictive algorithm, focusing entirely on the information properties of the data batches to visualize and identify temporal and spatial variations.
- The supervised module enables the direct measurement of AI model performance degradation across distinct temporal or sources batches by automating the training and evaluation of baseline predictive models.

This dual approach provides a comprehensive operational suite for addressing covariate, prior, and concept shifts in dynamic healthcare settings. Figure 1 describes the general workflow of the dashi package.

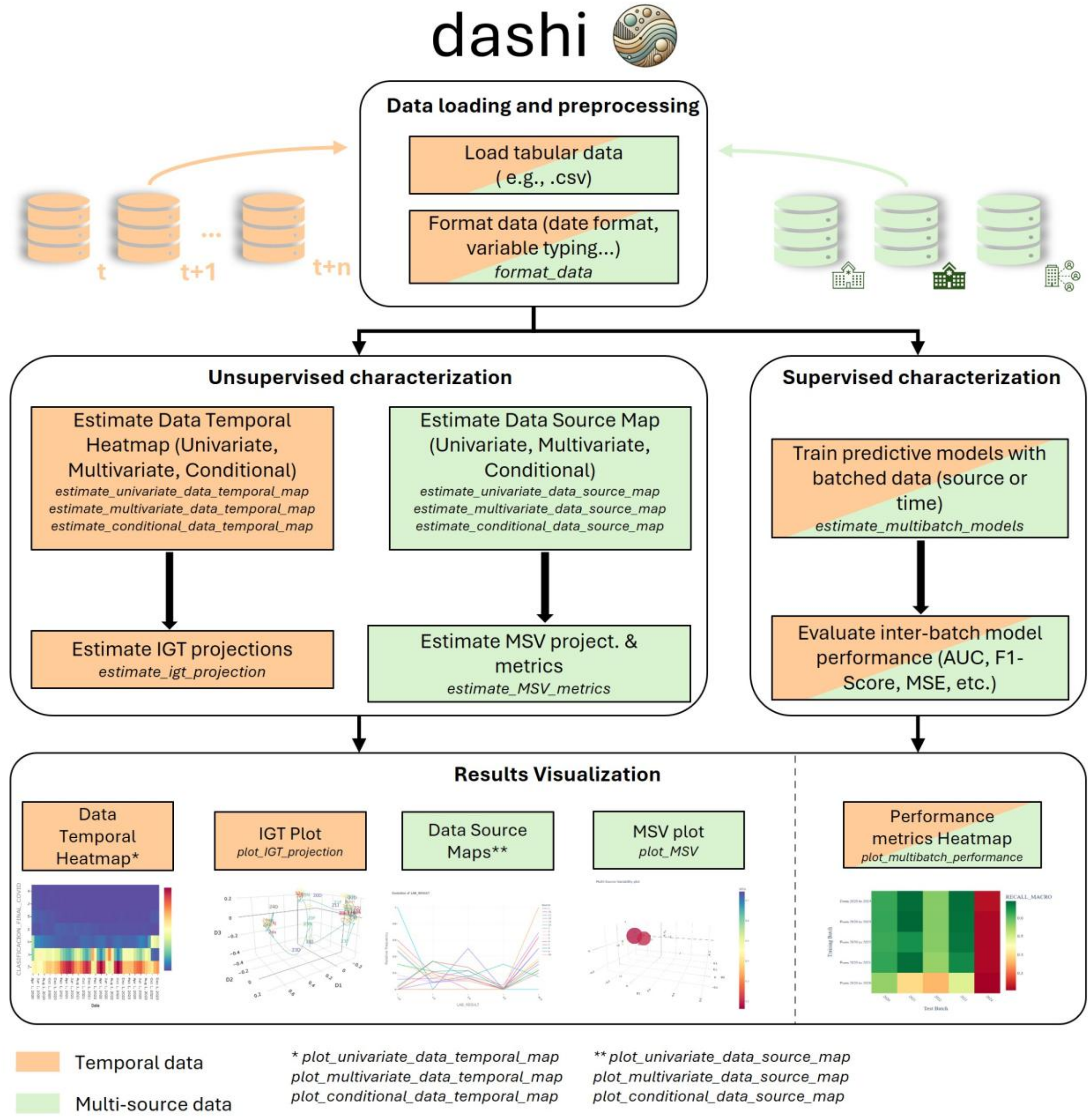

**Figure 1.** Workflow of the dashi Python package. Tabular data are loaded and formatted, then characterized in two ways depending on the nature of the data: (i) temporal structure leading to IGT projections, and (ii) source heterogeneity producing MSV projections and metrics. A supervised stage trains multi-batch models (by time or source) and evaluates inter-batch performance. Outputs include data temporal maps, IGT plots, data-source maps, MSV plots, and performance-metric heatmaps. The classification metrics include: AUC (per-class and macro), accuracy, log-loss, and precision/recall/F1 (per-class and macro/micro/weighted). Regression metrics: MAE, MSE, RMSE, and $R^2$. *IGT*: Information Geometric Temporal; *MSV*: Multi-Source Variability.

### 2.1 Data loading

To ensure immediate compatibility with the standard scientific Python ecosystem, dashi natively interfaces with fundamental data science libraries such as pandas and scikit-learn. This design choice allows practitioners to directly ingest tabular data, the standard format for clinical AI data feeds. Consequently, users of standardized EHRs, such as OMOP, i2b2, HL7-FHIR or OpenEHR, alongside operational logs, or epidemiological registries can convert the target inputs and outputs to standard pandas DataFrames, which are usable with dashi. Therefore, the library minimizes the friction typically associated with implementing advanced distribution shift analytics in production environments, thereby bridging the gap between theoretical data drift detection and practical health operations.

While the current native implementation of the dashi library is optimized for the direct ingestion of structured tabular data, its underlying mathematical framework remains fundamentally agnostic to the original data modality. In modern clinical environments, a significant amount of information is embedded within unstructured data, particularly medical images and free-text clinical narratives. To accommodate these complex data types, *dashi* can be seamlessly extended to analyze multimodal pipelines through intermediate preprocessing steps.

For unstructured clinical text, the integration process requires transforming raw data into continuous, high-dimensional numerical representations, commonly referred to as contextual embeddings. In the case of free-text data, it can be encoded leveraging modern Natural Language Processing models, such as transformer-based architectures fine-tuned on medical corpora. Once these contextual embeddings are extracted, they are ingested directly into the dashi pipeline as standard numerical features. Similarly, the medical images can be analysed by getting a fixed size representation in a latent space. This methodological bridging enables dashi to effectively track dataset shifts on multimodal data.

### 2.2 Unsupervised module

The unsupervised module of dashi provides a model-agnostic framework for evaluating dataset shifts by directly analyzing the probability distributions of features. The process estimates empirical distributions and quantifies their temporal and source-based divergence using the Jensen-Shannon (JS) distance. This information is then projected onto interpretable non-parametric statistical manifolds.

#### Distribution estimation: Univariate, Multivariate, and Conditional Analysis

For analyzing data evolution over time or across sites, dashi constructs different types of comprehensive maps, based on the probability distributions of each data batch over a shared support space: first, the data temporal maps, where the colour of the pixel at a given (X, Y) position indicates the frequency (absolute or relative) at which value Y was observed on date X [17]; and second, the data source maps, where the distinct trajectories represent the probability distribution across its domain of values (absolute or relative) of multiple sources. This foundational data to distribution mapping can be customized to suit varying levels of data complexity. For tracking features individually, the library generates univariate maps. To capture the complex, multivariate relationships inherent in data, dashi incorporates a multivariate analysis based on dimensionality reduction. To efficiently process high-dimensional and heterogeneous data, the library integrates robust dimensionality reduction techniques, such as Principal Component Analysis (PCA) and Singular Value Decomposition (SVD) for numerical data, Multiple Correspondence Analysis (MCA) for categorical data, and Factor Analysis of Mixed Data (FAMD) for mixed data. Following the projection of the high-dimensional data into a manageable lower-dimensional subspace, dashi computes a continuous joint probability density function using Kernel Density Estimation, evaluated across a discretized 2D or 3D meshgrid.

Although restricting the Kernel Density Estimation (KDE) to a maximum of three dimensions may entail a partial loss of original feature variance, this constraint is necessary to enable visual analytics. Because dashi prioritizes human-in-the-loop monitoring of clinical AI systems, visual interpretability is deliberately favored over lossless high-dimensional estimation. This ensures that complex temporal and multi-source dataset shifts remain intuitive and rapidly diagnosable by the users.

Finally, to construct the multivariate data temporal maps and data source maps, this resulting density tensor is marginalized across the relevant spatial dimensions, projecting the multidimensional probability distribution down into interpretable, one-dimensional density profiles.

Furthermore, to specifically investigate concept shifts and class-dependent variations, dashi enables conditional analysis. Instead of strictly evaluating the marginal distribution $P(X)$, the library can extract the conditional probability $P(X|Y)$, allowing the users to isolate and characterize how feature distributions evolve with respect to specific outcomes or target labels.

**Distribution Distance Metrics**

To quantify the divergence between temporal or spatial batches, dashi employs probabilistic distance metrics, primarily utilizing the JS divergence (Equation 1). For two distinct probability distributions $P$ and $Q$ (e.g., the distribution of a laboratory value in two different months), the JS divergence is formulated symmetrically based on the Kullback-Leibler divergence ($D_{KL}$) as:

$$\mathrm{JSD}(P \parallel Q) = \frac{1}{2} D_{\mathrm{KL}}(P \parallel M) + \frac{1}{2} D_{\mathrm{KL}}(Q \parallel M) \quad (1)$$

where $M = \frac{1}{2}(P + Q)$. The JS distance, defined as the square root of the JS divergence ($\sqrt{JSD(P \parallel Q)}$), satisfies the properties of a true mathematical metric. When calculated using base-2 logarithms, it is bounded between 0 and 1, providing a standardized and highly interpretable scale for comparing heterogeneous health distributions, where 0 means equal, overlapping distributions, and 1 means non-overlapping distributions.

**Information Geometric Temporal (IGT) Plots**

In the case of temporal shifts, dashi characterizes and visualizes the temporal evolution of these distributions via Information Geometric Temporal (IGT): a metric that measures the similarity between temporal data batches in a statistical manifold [14]. By calculating the pairwise JS distances between all sequential temporal batches of a dataset, the library constructs a comprehensive distance matrix. Classical Multidimensional Scaling (MDS), Non Metric MDS or PCA is subsequently applied to project this inter-batch dissimilarity matrix into a 2D or 3D non-parametric statistical manifold [18]. In an IGT plot, each coordinate point represents a distinct temporal batch, and the distance between points is directly related to their distribution similarity. The trajectory connecting these chronological points, calculated using splines between temporal points, empowers data scientists to visually track continuous or abrupt shifts and seasonality over time, as well as delineate temporal subgroups. These subgroups can be defined as conceptually related data batches at which probability distributions are similar within a group, but dissimilar between groups [18]. Most of the plots in the library include different color palettes, suited for different types of color-blindness, which improves visibility.

### Multi-Source Variability Plots and Metrics

To evaluate data variability across heterogeneous healthcare settings, dashi provides a multi-source characterization analysis structured around the Multi-Source Variability (MSV) space. Analogous to the previously described IGT methodology, this approach quantifies the divergence among source-specific probability distributions utilizing the JS distance. The resulting pairwise statistical distances are subsequently projected onto a non-parametric statistical manifold via MDS. To formally operationalize this multi-source variability, the projected MSV space is defined by two core metrics, explained below: Global Probabilistic Deviation (GPD) and Source Probabilistic Outlyingness (SPO) [19].

The GPD measures the degree of global multi-source variability. It computes a single, global scalar that quantifies the total dispersion of all data sources within the projected statistical manifold. Higher GPD values indicate a higher risk of multi-source dataset shift, serving as an automated warning that an AI model may suffer from severe calibration issues and poor generalization when trained and deployed on different settings.

The SPO measures the specific divergence of an individual data source relative to the central tendency of the remaining sources. In clinical practice, this metric can be critical for identifying aberrant batches or "outlier" hospitals that exhibit unique demographic distributions or localized coding practices, allowing AI training pipelines to be triggered before models fail silently on that specific sub-population.

## 2.3 Supervised module

The supervised module of dashi provides a model-centric methodology to directly measure the impact of dataset shifts into the predictive performance. This module is specifically designed to simulate real-world clinical AI evaluation workflows, answering the critical question of whether an observed distributional drift translates into a clinically significant "silent failure" of a predictive system.

### Automated Inter-Batch Training-Test Pipeline

To operationalize this assessment, the supervised module automates the generation and cross-batch evaluation of predictive models across the specified temporal or multi-source scopes. Natively, dashi utilizes robust, tree-based algorithms, specifically Random Forests and Histogram Gradient Boosting, which are well-suited for the non-linear relationships and heterogeneous feature types typical of real-world health data. These algorithms can be configured for both classification and regression tasks, depending on the outcome of interest. Furthermore, these tree-based models offer a highly cost-effective training phase while maintaining strong predictive performance on tabular datasets, efficiently capturing complex non-linear relationships. Although deep learning architectures currently represent the state-of-the-art in many predictive domains, their computational overhead and training costs remain substantial. Consequently, the current iteration of the dashi library prioritizes algorithmic usability, computational efficiency, and robust shift characterization over absolute predictive maximization. Future iterations of the library may encompass the implementation of deep learning models or provide modularity for users to integrate their own custom predictive architectures.

For each temporal or source-based batch, the data is partitioned into a standardized 80-20 training-testing split. Categorical variables are transformed using ordinal encoding for Histogram Gradient Boosting to leverage its native handling of categorical splits, whereas one-hot encoding is applied for Random Forests. Although tree-based ensembles are inherently scale-invariant, all continuous numerical features are automatically standardized via robust scaling. This design choice deliberately establishes a generalized, scalable preprocessing pipeline, ensuring the framework's architecture remains compatible with the future integration of scale-sensitive algorithms and avoiding data leakage. dashi allows users to define custom hyperparameter configurations for the predictive models. However, if no custom values are provided, the library defaults to a pre-defined set of conservative hyperparameters, chosen to balance performance, computation and usability.

The core evaluation pipeline operates through a batch-based training and testing methodology. The library offers two different training strategies depending on the training data selection: cumulative and from-scratch. The cumulative strategy trains the models using data accumulated from subsequent data batches, and it is designed for the temporal evaluation. In contrast, the from-scratch strategy trains each model using a single batch (e.g., a chronological block or data from a certain source). All trained models are then systematically evaluated against all temporal or multi-source test batches, generating a comprehensive matrix of cross-batch predictions.

**Model Performance Metrics**

dashi evaluates performance changes by tracking the confusion-matrix derived metrics across the defined batches with exploratory performance matrices. The performance metrics in dashi include AUC (per-class and macro), accuracy, log-loss and precision/recall/F1 (per-class and macro/micro/weighted) for classification tasks, and MAE, MSE, RMSE and $R^2$ for regression tasks.

By comparing the baseline validation performance against the performance on external data batches (temporal or source), practitioners can quantify the exact magnitude of performance drop-off caused by dataset shifts.

## 3. Results: Health AI case studies

To validate the operational utility of the dashi library, we present results of its application in three case studies in the field of health AI encompassing both simulated and real data.

### 3.1. Simulated Gestational Diabetes Mellitus (GDM) Cohort: Temporal Concept shift

For the first case study, we created a simulated GDM dataset comprising 96,000 patient records, collected in monthly batches between 2021 and 2024 from a single hospital (openly accessible in Supplementary Material 1). Grounded in a routine prenatal clinical workflow, the predictive task leverages Oral Glucose Tolerance Test (OGTT) results alongside standard clinical covariates (fasting and 1-hour glucose, HbA1c, Body Mass Index, and maternal age) to classify positive GDM cases. To emulate a realistic shifting situation, we introduced a transient data quality degradation driven by inconsistent coding practices, such as delayed or erroneous postpartum diagnoses. Consequently, during the initial two years of the study period, a substantial feature overlap between GDM-positive and GDM-negative classes is present. In statistical terms, the underlying clinical profiles of both groups become highly entangled within the feature space, effectively blurring the decision boundary for any predictive model and simulating a severe concept shift. Following simulated clinical audits and retraining initiatives, coding accuracy gradually improves throughout the third year, restoring strict label consistency by the fourth year. This case study was used in an external validation procedure during a workshop in the EFMI-STC 2025 conference, which counted with 25 participants, and led to several feedback used to improve the current version of dashi.

The unsupervised conditional analysis with dashi successfully isolates the underlying data anomaly. The generated conditional data temporal maps reveal that during the early months of the cohort, a substantial fraction of encounters labeled as GDM-positive paradoxically exhibit low fasting blood glucose values (Figure 2A), whereas the distribution for negative cases remains temporally stable (Figure 2B). This structural misclassification visibly resolves from late 2022 into early 2023. Furthermore, the IGT projection of the conditioned by class multivariate analysis (Figure 2C) captures this global evolution, clearly delineating two distinct temporal subgroups separated by a topological shift that perfectly aligns with the simulated auditing period. Finally, the supervised module evaluates the downstream impact of this instability; the cross-batch performance matrix (Figure 2D) demonstrates severe generalization failures when models trained on the initial, poorly coded batches are evaluated on later data, while overall predictive robustness significantly increases following the correction of the labeling protocols. A more detailed analysis of this case study can be found in Supplementary Material 2.

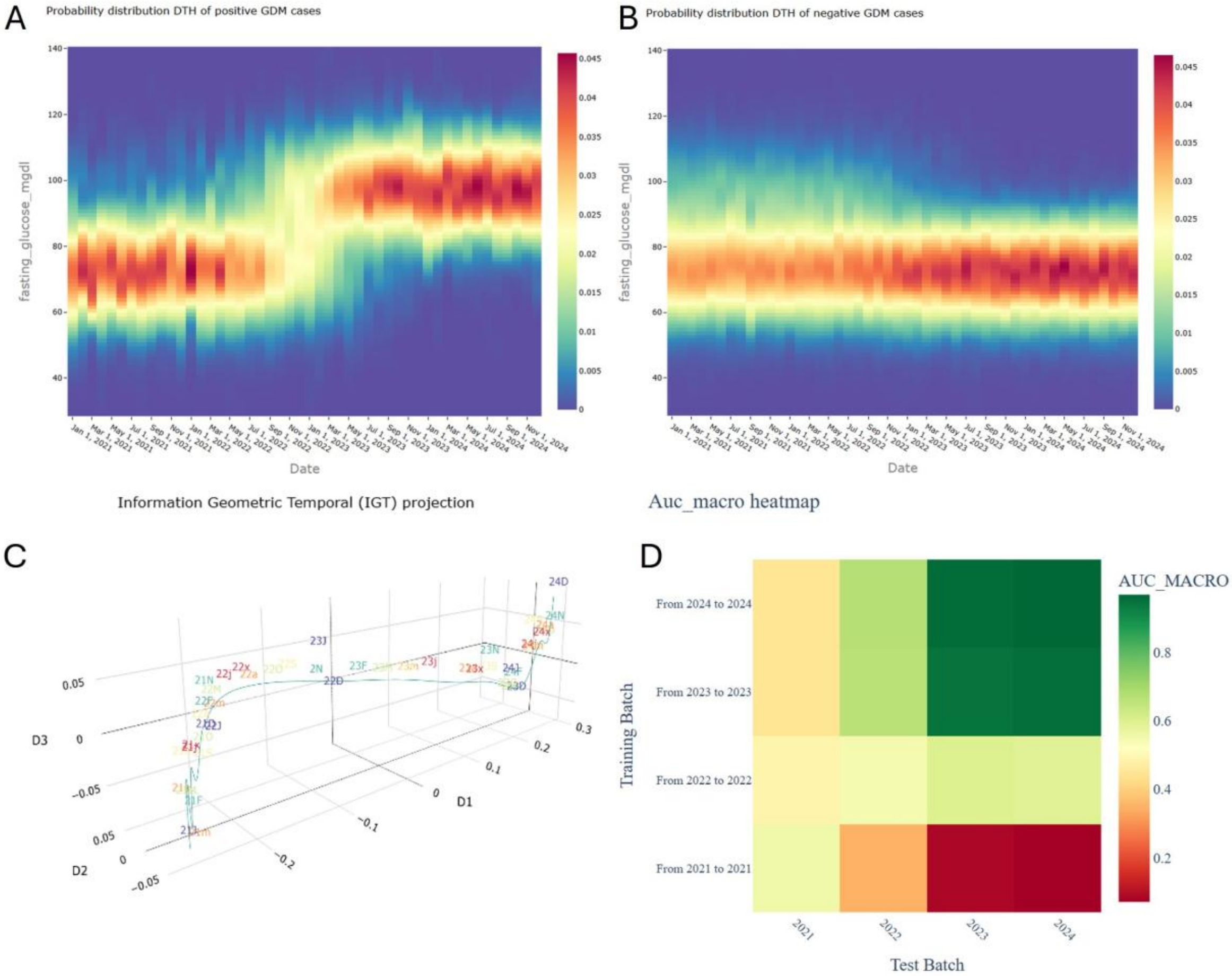

**Figure 2.** Characterization of temporal concept shift in the simulated Gestational Diabetes Mellitus (GDM) dataset using dashi. **(A, B)** Data temporal heatmaps illustrating the evolution of the fasting glucose probability distribution for positive (A) and negative (B) GDM cases. **(C)** Information Geometric Temporal (IGT) projection plot visualizing the trajectory of the multivariate conditional distribution over time. D1, D2 and D3 axis represent the projection of the components of variance. **(D)** Cross-time performance matrix displaying the MACRO-AUC of a Random Forest classifier, trained with from scratch strategy and evaluated across yearly train-test temporal batches to assess predictive model degradation.

### 3.2. Mexico COVID-19 Cohort: Multi-Source Domain Adaptation

The second case study demonstrates dashi's utility in characterizing multi-source variability and spatial domain shifts using the open real-world COVID-19 registry data from Mexico (2020 to 2024) [20] (Supplementary Material 3). For this experiment, we extracted a random subsample of 4,000 patient records per year and formulated a classification task predicting COVID-19 test positivity. To evaluate domain adaptation challenges across the healthcare network, the data was batched according to the specific type of National Health System (NHS) institution that provided patient care.

The unsupervised module effectively exposes severe structural heterogeneities across the institutional network. The distribution maps of the first three principal components from the multivariate analysis (Figure 3A) reveal distinct topological differences, particularly for institutions designated as NHS type 1 and NHS type 15. The MSV plot (Figure 3B) further contextualizes this disparity: while NHS types 12 and 4 process a high volume of data (represented by larger bubble sizes), NHS types 1, 2, and 5 suffer from severe underrepresentation. Crucially, the MSV projection isolates NHS types 1 and 15 as highly divergent from the central clinical consensus, presenting also the highest SPO values. Finally, the label analysis (Figure 3C) highlights significant prior shifts, contrasting NHS types with extreme class imbalances (e.g., types 99 and 15, where almost all cases belong to the positive class) against institutions with balanced prevalence rates (e.g., site 4).

To ensure robust model training, institutions with critical underrepresentation or extreme class imbalance were systematically excluded from the supervised evaluation. The resulting cross-institution performance matrix (Figure 3D) evaluates the predictive degradation when models are transferred between the remaining domains. Sources at the performance matrix are organised using a biclustering algorithm, a parametrised option in dashi. This structural reordering groups institutions based on mutual or group model transferability, revealing clusters of healthcare settings that share highly compatible underlying data distributions (yielding strong cross-batch performance) versus domains where uncalibrated model transfer results in significant performance drops. A more detailed analysis of this case study can be found in Supplementary Material 4.

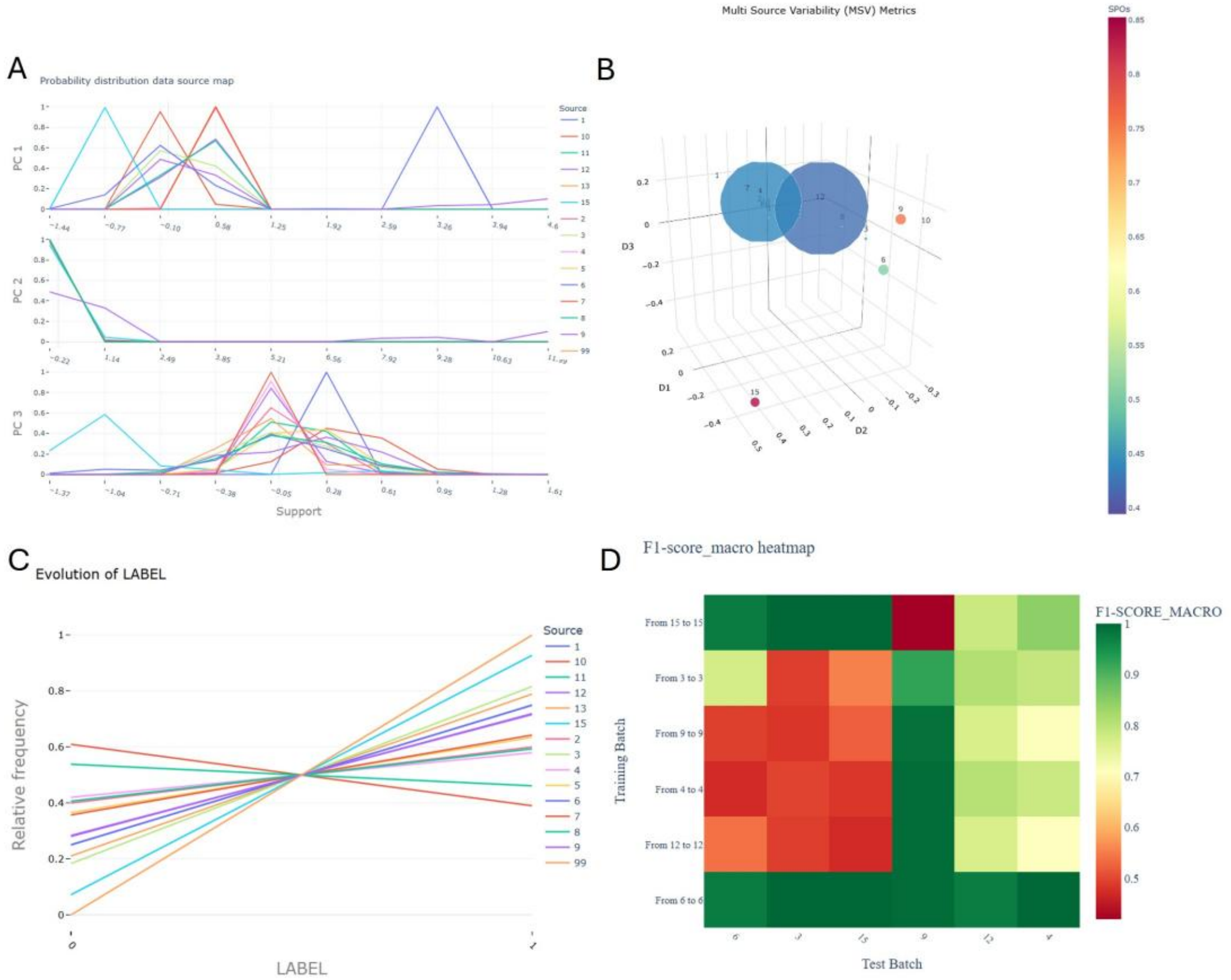

**Figure 3.** Characterization of multi-source variability in the Mexico COVID-19 dataset using dashi. **(A)** Multivariate data source map visualizing the distribution differences across sites utilizing the first three principal components of PCA. **(B)** Multi-source variability (MSV) metrics plot quantifying cross-site disparities. Distance between sites is determined through Global Probabilistic Deviation (GPD) and the color through Source Probabilistic Outlyingness (SPO). Size is determined by the amount of data in each site. **(C)** Univariate data source map illustrating the probability distribution of the binary target label across the respective sites. **(D)** Cross-site performance matrix displaying the macro-averaged F1-Score of a Random Forest classifier, demonstrating model degradation and domain adaptation challenges when trained and

### 3.3. Valencian Community 112 Emergency Medical Dispatch: Multimodal Temporal Shift

The final case study evaluates dashi's capacity to characterize unstructured, free text data dynamics using a large-scale, real-world registry of 2,574,230 Emergency Medical Call Incidents from the Valencian Community Health Services Department, spanning from 2014 to 2025. The data use was approved by the ethical committees of Fundació per al Foment de la Investigació Sanitària i Biomèdica de la Comunitat Valenciana (FISABIO) and Universitat Politècnica de València (UPV), in the context of the KINEMAI research project.Ferri et al., (2025)

The predictive task is formulated as a binary classification of the immediate life-threat risk associated with each incident, based entirely on the dispatcher's free-text clinical notes. As formalized in Section 2, integrating unstructured text into the dashi pipeline requires an initial preprocessing step. We transformed the raw clinical narratives into fixed-dimensional contextual embeddings utilizing the pretrained BioLord-2023 M model [22].

These embeddings were subsequently ingested into dashi as standard continuous variables for temporal characterization.

The unsupervised temporal analysis successfully maps the complex, long-term evolution of these emergency reports. The multivariate distribution heatmaps of the text embeddings (Figure 4A) identify an abrupt, severe structural alteration at the end of 2014, particularly visible in the second and third principal components. This shift is due to a change in the protocols and information storage system. This finding is consistent with the results of previous studies [23]. There is a secondary noticeable distributional change in March 2020, whose source is yet to be discovered. The IGT projection (Figure 4B) corroborates these findings, topologically isolating the pre-2014 batch from subsequent years. Furthermore, the IGT projection exposes a continuous longitudinal drift with a prominent shift in July 2021, and successfully captures strong seasonality, forming a topological axis for warm and cold months based solely on the latent linguistic patterns of the calls. This seasonality can be identified by the plot’s colours. The conditional multivariate mapping (Figure 4C) indicates only a slight divergence between the classes, as the conditional distributions $P(X|Y)$ are largely affected by the overall marginal dynamics shown in Figure 4A.

The supervised module's cross-batch performance matrix (Figure 4D), evaluated via ROC-AUC, translates these data-centric observations into operational insights. The severe covariate shift observed in 2014 causes profound model degradation; algorithms trained exclusively on this early temporal batch yield the lowest generalization scores across the timeline. Interestingly, while training on the entire cumulative dataset (top rows) yields stable performance on recent batches (right columns), the matrix reveals an anomalous peak in test performance for the year 2020, entirely independent of the training batch utilized. Retrospective clinical analysis indicates that the strict COVID-19 lockdown measures drastically reduced the volume of varied traumatic and routine emergencies, temporarily homogenizing the incident typology and simplifying the predictive decision boundary.

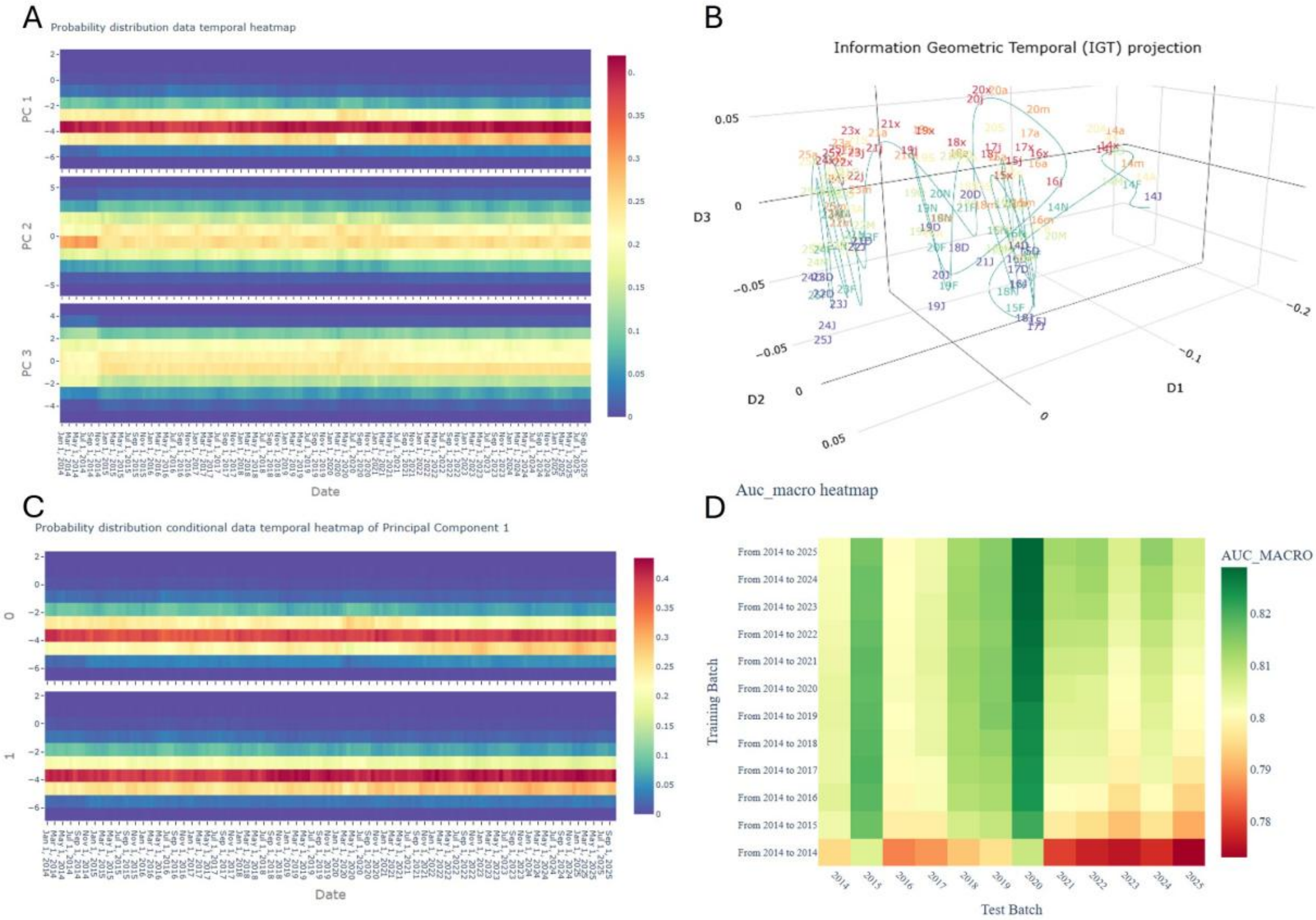

**Figure 4.** Characterization of 112 Emergency Calls dataset's temporal variability using dashi. **(A)** Multivariate data temporal heatmap displaying the evolution of contextual embeddings across the first three principal components following Singular Value Decomposition (SVD). **(B)** Information Geometric Temporal (IGT) projection plot characterizing the temporal variability of the dataset across monthly batches. **(C)** Conditional multivariate data temporal heatmap illustrating the evolution of the data distribution conditioned on the LIFE-THREAT target class over time. **(D)** Cross-time performance matrix presenting the MACRO-AUC of a Histogram Gradient Boosting classifier for LIFE-THREAT prediction, evaluated using a cumulative yearly batched training strategy.

## 4. Discussion

### Recommendations based on dashi outcomes

Table 1 relates a comprehensive taxonomy of the multi-source and temporal dataset shifts frequently encountered in data feeding AI, with effects on AI modeling and potential action strategies. Following a health AI rationale, where these shifts are common and of critical impact, we outline how the dashi Python library facilitates the identification of these phenomena, utilizing the different resources offered by the library, and details their specific impact on AI reliability, alongside recommended mitigation strategies. dashi can also be combined with the monitorization of calibration, Receiver Operating Characteristics (ROC) and Precision Recall (PR) curves, as well as fairness metrics, to build a comprehensive trustworthy-based AI framework [24].

**Table 1.** Recommendations based on dashi outcomes on dataset shifts in clinical AI: Characterization with dashi, impact on predictive modeling, and recommended mitigation strategies.

| Problem *[1] | Identification via dashi *[2] | Effect on AI Models | Possible Solutions & mitigation Strategies |
| --- | --- | --- | --- |

| (Shift typology) | | | |
|---|---|---|---|
| **Abrupt / Sudden Shifts**<br><br>(e.g., Pandemic onset, sudden clinical protocol changes, EHR system migrations) | **Data Temporal Maps:** Abrupt changes in data distributions in short time periods (Figure 4A).<br><br>**IGT Plots:** Rapid increase in the distance between subsequent temporal batches (Figure 4B).<br><br>**Supervised metrics:** Rapid drop in the performance of the tracked metrics between subsequent temporal batches (Figure 4D). | **Development phase** | |
| | | If present in retrospective datasets, abrupt shifts may split the data into heterogeneous regimes. Training and validation subsets drawn across these regimes can lead to misleading performance estimates and, if unmanaged, potentially biased delivered models. | Split or stratify the cohort around the detected event; train separate pre-/post-shift models or prioritize post-shift data (e.g., using continual learning strategies) depending on the deployment data; document the event as a dataset-version boundary. The case of punctual abrupt shifts can be treated as outliers, discarding those batches. |
| | | **Deployment phase** | |
| | | When a shift occurs after deployment, models may face data that no longer follows the assumptions learned during training. This can result in sudden degradation in predictive performance and risk to patient safety. | Trigger urgent review, recalibration of precision thresholds or retraining depending on whether new data is available; apply post-shift instance weighting; temporarily restrict or freeze automated CDSS use until safety is re-established. |
| **Gradual Shifts / Trends**<br><br>(e.g., Slow demographic aging, gradual evolution of medical guidelines) | **Data Temporal Maps:** Continuous, progressive changes in the probability distribution over temporal batches (Figure 2A).<br><br>**IGT Plots:** Observation of continuous linear or polynomial temporal trajectories (Figure 2B).<br><br>**Supervised metrics:** Continuous and gradual drop in the performance of the tracked metrics in prospective performance. | **Development phase** | |
| | | Historical data may not be equally informative for the prospective target period. Conventional random splits can overestimate robustness by mixing old and recent batches. | Use temporally aware training and validation, using continual learning approaches such as sliding-window or time-weighted training, and sensitivity analyses by epoch; remove or redesign highly time- dependent variables. |
| | | **Deployment phase** | |
| | | Slow model obsolescence: performance and calibration degrade gradually and may remain unnoticed until a model revision is done. | Schedule periodic retraining and recalibration; define trend-based monitoring thresholds to trigger maintenance before the model becomes unsafe. Adopt automated continual re-training strategies, always under data protection and AI regulations. |
| **Seasonality**<br><br>(e.g., Flu season incidence, temperature-dependent respiratory exacerbations) | **Data Temporal Maps:** Periodic fluctuations in probability distribution.<br><br>**IGT Plots:** In monthly and weekly analysis it can be identified by grouped similar colours at a transversal direction over the layout of batches (Figure 4B). It can also be observed as cycles in temporal trajectories.<br><br>**Supervised metrics:** Seasonal variation in the performance of the tracked metrics. | **Development phase** | |
| | | A cohort dominated by one season may underrepresent relevant information from other seasons, biasing feature importance and validation estimates. | Ensure balanced seasonal sampling; include calendar or seasonal features when clinically justified; evaluate models with season-stratified metrics. |
| | | **Deployment phase** | |
| | | Seasonal patterns can cause recurring fluctuations in model performance as incoming data distributions change periodically, with reliability varying according to how well each season is represented during development. | Use season-aware recalibration, monitoring thresholds, or ensembles with season-specific sub-models when recurrent seasonal degradation is observed. |
| **Temporal Subgroups**<br><br>(typically stemming from abrupt / sudden shifts separating distinct periods such as Pre-intervention vs. Post-intervention) | **Data Temporal Maps:** Different temporary-grouped patterns in data distributions of batches over the full period (Figure 2A).<br><br>**IGT Plots:** Identifying distinct, isolated clusters of | **Development phase** | |
| | | Temporal subgroups indicate distinct data regimes. Ignoring them during development can mask period-specific weaknesses, inflate robustness estimates, and | Use epoch-stratified validation; train pooled, subgroups-specific or invariant models depending on transferability; report performance separately by subgroup. |
| | temporal batches (Figure 4B).<br><br>**Supervised metrics:** Assessing severe off-diagonal performance drops in inter-epoch performance matrices (Figure 1D). | produce uneven performance across subgroups. | |
| | | **Deployment phase** | |
| | | Deployment into a different temporal subgroup can alter model behaviour, affecting prediction reliability, calibration, and error distribution even when aggregate metrics remain acceptable over the whole period. | Detect subgroup transitions as monitoring events; switch to a subgroup-specific model, recalibrate, or require revalidation before unrestricted deployment. |
| **Multi-Source Variability**<br><br>(e.g., Deploying a model developed at one hospital or group of hospitals to another distinct healthcare center) | **Data Source Maps:** Differences in the probability distribution between the sources (Figure 3A).<br><br>**MSV plots & metrics:** Grouped or isolated sources (e.g., sites). High GPD across sources or high SPO for outlying sources (Figure 3B).<br><br>**Supervised metrics:** Drop in performance when cross-evaluating models between distinct sources, either grouped performances using the bi-clustering mode or at specific sources (Figure 3D). | **Development phase** | |
| | | In multi-site training, ignoring heterogeneous source-specific distributions may overestimate generalizability, mask source-specific weaknesses, and produce uneven performance across sources. | Evaluate source overlap before pooling data; use source-aware validation, stratified sampling, harmonization, and domain adaptation strategies such as instance weighting, batch effect correction, feature alignment, transfer learning... |
| | | **Deployment phase** | |
| | | Deployment in a source with a different distribution can alter model behaviour, affecting prediction reliability, calibration, and error distribution, especially when monitoring is not stratified by source. | Monitor each source separately; apply local recalibration, source-specific fine-tuning, transfer learning, or domain adaptation for divergent sources, and restrict deployment when cross-source transportability is insufficient. |

*[1] The shift typologies can occur either at covariate shift, concept shift and prior shift level.

*[2] dashi's methods are capable of diagnose both covariate and concept shifts. While covariate and concept shifts are normally mutually related, the dashi's Conditional analysis route (in Data Temporal Maps, IGT Plots and MSV methods) can help detect concept shift with a higher gain in cases where the changes occur differently at the distinct outcomes to be predicted: e.g., in sepsis prediction label definition changes to a new criteria (e.g., SIRS to SOFA) changing the $p(x|y)$.

## Significance

The critical value of dashi lies in its ability to support model robustness, safety and interpretability against real-world shifting environments, in a cost-effective manner. It provides an accessible, end-to-end Python toolkit specifically designed to translate the complex mathematical theory of dataset shifts into actionable, day-to-day operations and pipelines. As demonstrated across our case studies, the assumption of stable data distributions in clinical environments is frequently violated, leading to severe model degradation. While theoretical machine learning literature extensively categorizes these phenomena, such as the spatial domain shifts observed in the multi-source Mexico COVID-19 cohort or the temporal covariate shifts seen in the 112 Emergency dataset, translating this theoretical awareness into routine AI lifecycle operations has remained a fundamental bottleneck.

Our findings prove that silent failures in CDSS are not merely theoretical risks, but observable and preventable events. By explicitly quantifying how an unnotified change in clinical coding (as seen in the simulated GDM cohort) or a sudden epidemiological lockdown (as seen in the 112 cohort) directly correlates with changes in data distributions or cross-batch performance degradation, dashi bridges the gap between data science theory and practical clinical safety. The findings from these diverse case studies underscore the necessity of managing dataset shift not as an isolated post-deployment patch, but as a continuous requirement across the entire AI lifecycle. By bridging the critical phases of model development, evaluation, and active deployment, the dual unsupervised and supervised approach presented in dashi facilitates rigorous algorithmic auditing at every stage. During initial development and evaluation, this approach proved essential for exposing latent cohort biases and quantifying temporal or multi-source variability before models ever reached the deployment phase. Furthermore, as models transition into operational deployment, this continuous monitoring acts as a vital early warning system. By detecting underlying data instabilities before they manifest silent clinical failures, healthcare institutions are empowered to proactively trigger model recalibration or restrict spatial domain transfer, thereby averting predictive obsolescence. Ultimately, this paradigm shift, moving from static model validation to continuous distributional surveillance, is a fundamental prerequisite for the engineering of robust, secure, and Trustworthy AI systems capable of operating safely within dynamic, real-world healthcare environments.

Dataset shifts may arise from underlying data quality problems, such as systematic biases in data collection, coding, or measurement practices, as well as changes in data quality dimensions including completeness, conformance, plausibility, or consistency [25]. Although traditional clinical data quality assessments remain necessary for basic database maintenance, their reliance on predefined rules and mostly univariate checks can limit their ability to detect the multivariate “unknown unknowns” characteristic of clinically relevant dataset shifts. In response to the limitations of static data quality checks, several open-source MLOps frameworks, most notably Evidently AI and NannyML [26], have emerged as commercial tools for automated post-deployment monitoring. While these tools provide valuable longitudinal tracking, their methodological approaches differ significantly from dashi. Tools like Evidently AI fundamentally evaluate data drift through a pairwise paradigm, comparing a single "current" data batch against a predefined "reference" baseline using standard, independent statistical tests (e.g., Kolmogorov-Smirnov or Chi-Square applied column-by-column). While effective for basic univariate drift, this approach struggles to capture the complex, non-linear relationships and joint distributional changes inherent in heterogeneous environments. Furthermore, because these evaluations are strictly pairwise, they do not inherently model the global topological structure of the entire time series.

The library dashi advances beyond these standard dataset shift detectors. Rather than relying on traditional, independent statistical tests, which predominantly evaluate data drift through isolated, univariate comparisons, dashi projects the entire joint probability distribution onto non-parametric statistical manifolds via IGT and MSV projections. This global topological approach captures complex, non-linear feature interactions, enabling the automated detection and comprehensive characterization of subtle and structural shifts, followed by quality control methodologies [17]. While frameworks like Evidently AI and NannyML are optimized for tracking a single deployment pipeline over time, dashi’s native multi-source metrics (GPD and SPO) provide a dedicated multi-source variability characterization framework, allowing practitioners to quantify and interpret shift across dozens of disparate hospital networks simultaneously.

**Limitations and future work**

While dashi provides a comprehensive framework for characterizing dataset shifts, its current implementation presents certain limitations, primarily concerning computational scalability with highly dimensional clinical data. The unsupervised module relies on the calculation of pairwise probability distances and the subsequent projection of these metrics onto non-parametric statistical manifolds. As the feature space expands linearly, the computational overhead required to construct these topological spaces increases exponentially. Although the library currently mitigates this through the integration of dimensionality reduction, natively processing massive, ultra-high-dimensional datasets, such as raw genomic, remains a computational bottleneck for real-time characterization. Future optimizations will focus on implementing approximate distance estimations and parallelized processing to handle these complex modalities more efficiently.

A second avenue of ongoing development is the native integration of multimodal data pipelines for free text and image data. As demonstrated in the Valencian Community 112 Health Service Department case study (Section 3.3), dashi is entirely capable of evaluating unstructured clinical text; however, this currently requires users to independently preprocess the text into contextual embeddings using external Natural Language Processing models before ingestion. Future iterations of the library aim to incorporate native, automated embedding pipelines.

A third area for future development involves expanding the supervised module's models. Future iterations of dashi are planned to natively integrate deep learning models. Ultimately, the library will be extended to allow the users introduce their own custom models into the evaluation pipeline.

Finally, future research will focus on transitioning dashi from a passive monitoring toolkit into an active MLOps controller. Currently, the library excels at visualizing shift and quantifying predictive degradation, but it relies on the practitioner to interpret these outputs and manually intervene. By conducting extensive retrospective analyses across diverse clinical cohorts, we aim to establish standardized, evidence-based actuation thresholds for the dashi's outcomes. Establishing these statistical cut-offs will enable dashi to automatically guide operational directives, such as triggering clinical risk flags or generating automated model retraining suggestions.

## 5. Conclusions

Systems based on machine learning techniques frequently face a major challenge when deployed in real-world environments: the conditions under which a system was developed rarely match the conditions of its eventual use. This fundamental problem is formally known as dataset shift. While traditional predictive models assume static environments, real-world data is inherently non-stationary, making robustness to shifting distributions a primary bottleneck for achieving reliable artificial intelligence.

As healthcare systems increasingly rely on predictive algorithms, the fundamental challenge for medical informatics has shifted from achieving initial state-of-the-art performance to guaranteeing longitudinal patient safety. dashi addresses this modern reality not merely as an analytical utility, but as a foundational mechanism for continuous algorithmic auditing.

By operationalizing complex statistical phenomena, such as temporal and multi-source topological variations, into a standardized, programmatic pipeline, this library lowers the barrier to ensure secure AI development and maintenance. It empowers healthcare institutions to transition from reactive analyses of silent clinical failures to a posture of proactive, real-time data governance.

Ultimately, dashi provides the necessary infrastructure to comply with emerging regulatory standards for Trustworthy AI. Ensuring that a model's underlying statistical assumptions remain continuously aligned with an ever-evolving clinical reality is the only sustainable path to safely scale artificial intelligence in the ever-evolving health ecosystem.

### Availability of source code and requirements

- Project name: dashi
- Project home page: https://dashi.upv.es
- Main Source Code repository: https://github.com/bdslab-upv/dashi
- Documentation: https://bdslab-upv.github.io/dashi/
- PyPi repository: https://pypi.org/project/dashi/
- Operating system(s): Platform independent.
- Programming language: Python 3.10 or higher.
- License: Apache 2.0

### Acknowledgements

Funded by Agencia Estatal de Investigación—Proyectos de Generación de Conocimiento 2022, project KINEMAI (PID2022-138636OA-I00).

### Competing interest

The authors declare that they have no competing interests.